\documentclass{article}

\usepackage{arxiv}

\usepackage[utf8]{inputenc} 
\usepackage[T1]{fontenc}    
\usepackage{hyperref}       
\usepackage{url}            
\usepackage{booktabs}       
\usepackage{amsfonts}       
\usepackage{nicefrac}       
\usepackage{microtype}      
\usepackage{lipsum}
\usepackage{graphicx}
\usepackage{amsmath}
\graphicspath{ {./images/} }

\title{Deep Metric Learning Model for Imbalanced Fault Diagnosis}

\author{
 Xingtai Gui \\
  University of Electronic Science and Technology of China \\
  \texttt{tabgui@163.com} \\
   \And
 Jiyang Zhang \\
  University of Electronic Science and Technology of China \\

}

\begin{document}
\maketitle
\begin{abstract}
Intelligent diagnosis method based on data-driven and deep learning is an attractive and meaningful field in recent years. However, in practical application scenarios, the imbalance of time-series fault is an urgent problem to be solved. This paper proposes a novel deep metric learning model, where imbalanced fault data and a quadruplet data pair design manner are considered. Based on such data pair, a quadruplet loss function which takes into account the inter-class distance and the intra-class data distribution are proposed. This quadruplet loss pays special attention to imbalanced sample pair. The reasonable combination of quadruplet loss and softmax loss function can reduce the impact of imbalance. Experiment results on two open-source datasets show that the proposed method can effectively and robustly improve the performance of imbalanced fault diagnosis.
\end{abstract}


\section{Introduction}
With the rapid development of modern industry and intelligent manufacturing, fault diagnosis of key equipment and production processes in smart factory is becoming more important for production safety and economic benefits 
\cite{jiao2020double,wang2018distribution}.
To ensure that faults can be diagnosed accurately and timely, it is necessary to mine useful information from large number of sensor signals. In recent years, data-driven methods have been extensively studied since they can perform effective process monitoring and fault diagnosis in automated equipment or systems where it is difficult to build a specific model
\cite{ge2017review, sun2021survey}.

Deep learning is widely used in the fields of computer vision, natural language processing, biological information \cite{I1,I2,N1,N2,B1} and so on because it can process a large amount of data and obtain effective features from these data. As sensing technology and data storage technology promoting, massive data are recorded in production process. Therefore, in the field of fault diagnosis, many classical and effective models in deep learning were verified to have good performance such as Deep Belief Network (DBN) \cite{DBN1,DBN2,DBN3}, Autoencoders (AE) \cite{AE1,AE2,AE3}, Convolutional Neural Network (CNN) \cite{CNN1,CNN2,CNN3}. Considering the fault signals are not independent of each other, recurrent neural network (RNN)-based methods were proposed and achieve performance improvement on dynamic time-series fault signals. Yuan et al. \cite{D4} used standard LSTM methods to predict the remaining life of aero engine in complex and strong noise interference environments. Cabrera et al. \cite{D5} iteratively trained a group of LSTM models from a time series representation and the hyperparameter search is guided by a Bayesian approach. Lu et al. \cite{D6} used LSTM to make use of previous context information and made the distribution estimator more robust to warp along the time axis.

Although deep learning methods have achieved certain accomplishment in fault diagnosis task, there still exits some challenges. As is known to all, sufficient high-quality data is the primary requirement to ensure the performance of deep learning-based fault diagnosis methods. When it comes to class imbalance, models will typically over-classify the majority group due to its higher prior probability. As a result, the instances belonging to the minority group are misclassified more often than those belonging to the majority group 
\cite{im1}.
However, during practical production process, it is difficult to collect sufficient and balanced data because it is time-consuming and costly to obtain sufficient fault samples and the occurrence of each fault is also different. Current methods dealing with imbalanced data are often based on oversampling 
\cite{im2}, but these methods do not provide additional information for the model. With the development of generative model, GAN is widely used in fault diagnosis and able to deal with imbalanced scenarios \cite{gan1, gan2, gan3}. However, the most methods based on data generation are two-stage.

To better model the edge and distribution of imbalanced class, deep metric learning (DML) has been applied in imbalanced data classification 
\cite{DML1,DML2}. Classical DML model such as siamese network \cite{siam} project data into a new metric space where the distance between different classes increases, and the dispersion within a class decreases. The triplet network inspired by siamese network contains three objects, called anchor, positive sample and negative sample 
\cite{trip}. Triplet network simultaneously considers intra-class and inter-class relationships to provide higher discrimination capabilities. Recently, DML-based model for fault diagnosis has been researched. Li et al. 
\cite{DML3} proposed a representation clustering algorithm to minimize the distance of intra-class variations and maximize the distance of inter-class variations simultaneously.  Wang et al. 
\cite{DML4} introduced a novel loss function called normalized softmax loss with adaptive angle margin (NSL-AAM) based on DML which can supervise neural networks learning imbalanced data without altering the original data distribution.

DML consists of three main parts which are informative input samples, the structure of network and  metric loss 
\cite{DML5}. Although imbalanced fault diagnosis model based on DML has been researched, there is no optimization strategy considering input sample pair. DML, as pair-based model, by adding a \textit{‘minor sample’} selected from imbalanced class in data pair besides positive sample and negative sample, will pay attention to the distance between anchor and positive, negative sample, imbalanced class data simultaneously. Based on such data pair, a quadruplet loss function referring to contrastive loss can be designed. By adjusting weight, the model pays more attention to the compactness of balanced class and distance between the imbalanced class and other classes. Making use of the powerful feature learning ability of deep learning, the distance between imbalanced and balanced class is as large as possible in the new mapping space, while maintaining the density of balanced class.

Based on the inspiration of above research, this paper proposed a quadruplet deep metric learning time-series fault diagnosis model (LSTM-QDM) under the condition of class imbalance. This model utilizes LSTM to extract feature from time series data and embeds the extracted feature into a low-dimensional space where the distance of intra-class decreases and the distance of inter-class increases. Simultaneously, this model makes the distance between imbalanced class and other classes even bigger and balanced class more compact to reduce the effect of imbalance.

The main contributions of this paper are summarized as follows.
\begin{itemize}
\itemsep=0pt
\item Based on traditional DML model, propose a new data pair design manner. In addition to positive and negative samples, select sample from imbalanced classes to increase the attention to imbalanced class. And the data pair initialization process considers single-class and multi-class imbalance scenarios simultaneously.

\item A quadruplet loss based on new data pair is developed to optimize deep model. The quadruplet loss considers distance relationship of anchor with postive, negative and minor samples. By setting discriminative weight for  data pair,  pays special attention on imbalanced class.

\item Design a new loss function combined softmax loss and quadruplet loss for imbalanced classification and conduct diverse imbalanced fault diagnosis experiments on Tennessee East-man process dataset and CWRU bearing fault dataset to prove the robustness and effectiveness of proposed method.
\end{itemize}

\section{Theoretical background}

\subsection{Recurrent neural network and long short-term memory network}
\label{lstm}
Recurrent neural network (RNN) 
\cite{RNN} adds a memory function compared to traditional forward neural network, so that the information of the neuron at the current moment contains the information of previous moment. RNN model can capture the correlation information between the input time series data. The forward propagation of RNN at time $ t $ are as follows.
\begin{equation}
\begin{aligned}
& h_t=\ \sigma(W_{hx}x_t+\ W_{hh}h_{t-1})\\
& o_t=\ f(W_{oh}h_t+b_0)
\end{aligned}
\end{equation}
Where $ x_t\in R^{N_i} $, $ h_t\in R^{N_h} $, $ o_t\ \in R^{N_o} $ are respectively the input vector, the hidden state vector and output vector. $ W_{hx}\in R^{N_h\ \times N_i} $, $ W_{hh}\in R^{N_h\ \times N_h} $, $ W_{oh}\in R^{N_o\ \times N_h} $ are respectively the weight matrix of input $ x_t $ and hidden state $ h_t $, weight matrix of previous time hidden state $ h_{t-1} $ and current time hidden state $ h_t $ and weight matrix of hidden state $ h_t $ and output $ o_t $. Function $ \sigma(\bullet) $ and $ f(\bullet) $ are respectively non-linear activation function of hidden layer and output layer. The hidden state at time $ t $, $ h_t $ is affected by hidden state $ h_{t-1} $ and input data $ x_t $. In the training process, three weight matrices, $ W_{hx} $, $W_{hh} $ and $ W_{oh} $, are updated with back propagation iteratively. During the optimization, RNN often has the risk of gradient vanishing and gradient exploding so it is unable to capture long-term sequence information \cite{RNN2}.

In order to improve these problem, long short-term memory (LSTM) was proposed on the basis of RNN
\cite{D3}. Compared with RNN, the hidden state in LSTM is controlled by complex control gate signal and memory signal to optionally forget or update time series information. The forward propagation of LSTM at time t are as follows.
\begin{equation}
\begin{aligned}
& f_t=\ \sigma(W_{fh}h_{t-1}+W_{fx}x_t)\\
& r_t=\ \sigma(W_{rh}h_{t-1}+W_{rx}x_t)\\
& o_t=\ \sigma(W_{oh}h_{t-1}+W_{ox}x_t)\\
& \widehat{c_t}=\ tanh(W_{ch}h_{t-1}+W_{cx}x_t)\\
& c_t=f_t\odot c_{t-1}+r_t\odot\widehat{c_t}\\
& h_t=o_t\odot\tanh{\left(c_t\right)}
\end{aligned}
\end{equation}

Where $ x_t $ and $ h_t $ are same as those in RNN, and the cell state,$ c_t $ is added in the model which is applied to store long-time information. $ f_t $, $ r_t $ and $ o_t $ are respectively forget gate, update gate and output gate. These gate signals are calculated with different weight matrix, current input vector and last time hidden state vector and embedded into interval of (0,1) to reflect the gates’ state.  $ \widehat{c_t} $ represents temporary memory information at time t and $ c_t $ is obtained by forgetting $ c_{t-1} $ and updating $ \widehat{c_t} $ to some degree. $ \odot $ represents Hadamard product operator. Finally, the hidden state at time $ t $ is calculated by output gate and $ c_t $. The activation function used in LSTM, $ \sigma $ and tanh, are sigmoid function and hyperbolic tangent function respectively.

\subsection{Deep metric learning}
\begin{figure}[h]
	\centering
		\includegraphics[scale=.6]{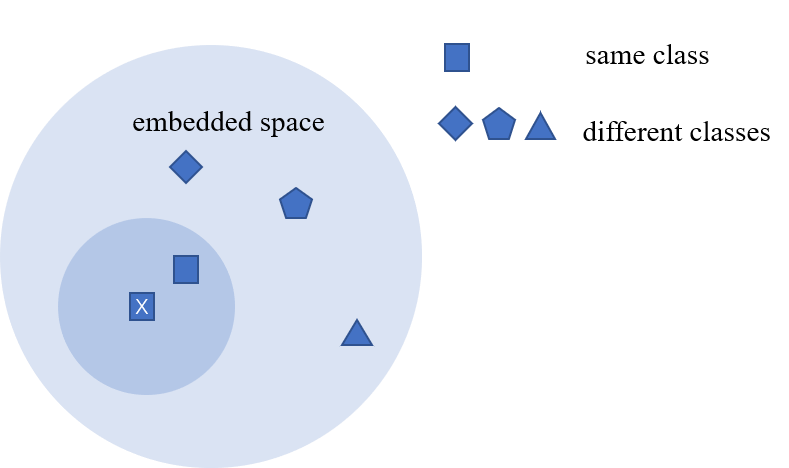}
	\caption{Ideal distribution in embedding space}
	\label{img2}
\end{figure}

Deep metric learning (DML) is a method of metric learning
\cite{DML5}, and its goal is to learn a map from original feature space to a low-dimensional dense embedding space where the samples from same class are closer, and the distance between different classes is farther. The distance is usually calculated by distance function such as Euclidean distance or Cosine distance. In embedding space, the ideal data distribution is shown in Figure \ref{img2}. Loss function plays an important role in DML and many of them is designed based on data pairs.

\begin{figure}[h]
	\centering
		\includegraphics[scale=.28]{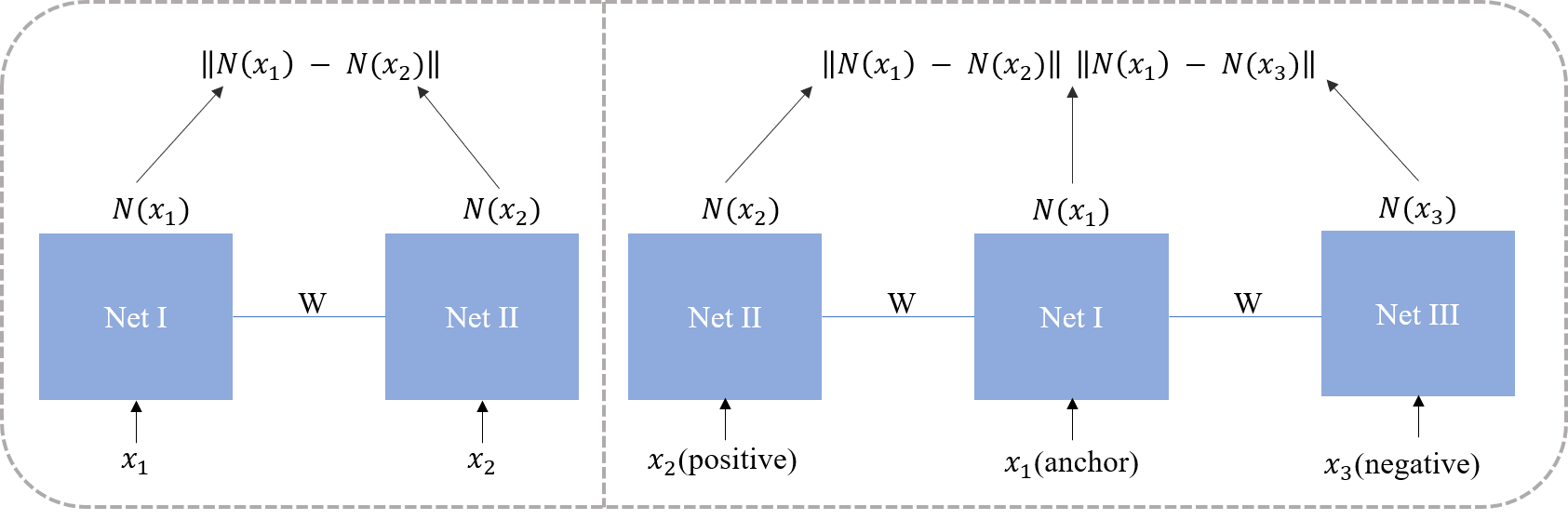}
	\caption{Structure of Siamese Network(left) and Triplet Network(right)}
	\label{img3}
\end{figure}

Siamese network is a DML model widely used in different fileds 
\cite{s1,s2,s3}. Siamese network is a symmetrical network consisting of two identical neural networks in parallel, whose structure and parameters are shared. It maps two input data into embedding space to obtain feature vector, and then uses the distance function to measure the similarity. As shown in Figure \ref{img3} left, $ x_1 $ and $ x_2 $ are fed into the symmetrical network and the loss function called contrastive loss is shown in Eq.\ref{siame}. $ Y $ represents whether the two input data are from same class and equals 1 if from same class and 0 if from different classes.
\begin{equation}
\label{siame}
\begin{split}
\begin{aligned}
& D = \left \| N(x_1)-N(x_2) \right \|_2 \\
& L =  YD +  (1-Y)max(0,margin-D)
\end{aligned}
\end{split}
\end{equation}

Triplet network is also applied widely
\cite{t1,t2,t3}. Similar to the siamese network, it consists of three identical neural networks in parallel, sharing all parameters as shown in the Figure \ref{img3} right, where $ x_1 $ from class C is called anchor and $ x_2 $, $ x_3 $ come from class C and other classes respectively called positive sample and negative sample. The loss function of triplet network is shown in Eq.\ref{triplet}. Generally, triplet network has a better performance than siamese network because it considers the relationship between anchor, positive sample and negative sample simultaneously.

\begin{equation}
\label{triplet}
\begin{split}
\begin{aligned}
& D_{pos} = \left \| N(x_1)-N(x_2) \right \|_2^2 \\
& D_{neg} = \left \| N(x_1)-N(x_3) \right \|_2^2 \\
& L = max(0, D_{pos} - D_{neg} + margin)
\end{aligned}
\end{split}
\end{equation}

\section{Time series fault diagnosis based on DML}
\begin{figure*}[h]
	\centering
		\includegraphics[scale=.6]{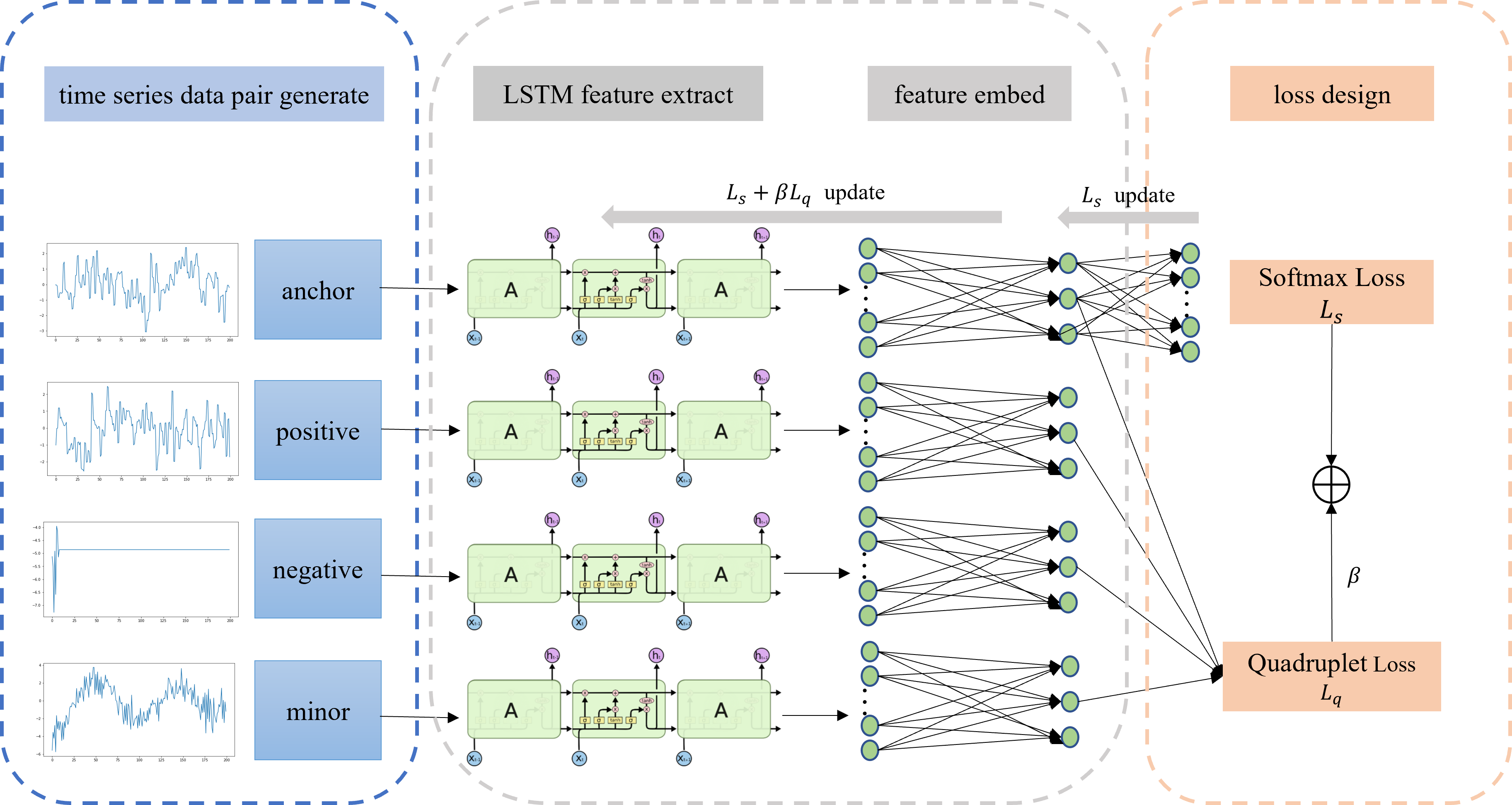}
	\caption{Structure of LSTM-QDM}
	\label{img4}
\end{figure*}

In actual production process, imbalanced fault diagnosis data exits commonly which make it difficult to build an accurate fault classifier based on traditional deep learning model. This paper proposed LSTM-quadruplet deep metric model (LSTM-QDM) based on the powerful time series feature extraction capability of LSTM and the ability of DML to adjust the distance between classes and within a class. The whole architecture of LSTM-QDM is shown in Figure \ref{img4}. The proposed model consists of three main parts: data pair design and generate, feature extract and embed, loss function design.

\subsection{Quadruplet time series data pair design}
\label{31}
To provide more attention to imbalanced class samples, referring to the data pair design of triplet network, in addition to anchor sample, positive sample and negative sample, minor sample is proposed based on imbalanced classes. As shown in Figure \ref{img4} blue part, positive sample is from the same class as anchor and negative is from the different and balanced classes, while minor sample is from imbalanced classes. QDM model considers scenarios of one-fault imbalance and multi-fault imbalance. When there exits one-imbalanced class, if anchor's class is imbalanced, then minor sample is randomly selected from negative classes while in multi-imbalanced classes condition, if anchor’s class is imbalanced, then minor sample is selected from a different class in imbalanced classes subset. The data pair consists of four sample including anchor is shown in Eq.\ref{pair}.
\begin{equation}
\label{pair}
\begin{aligned}
& Pair_n=(x_n,\ x_{pos},\ x_{neg},\ x_{minor})\\
& x_{pos}\in\left \{ C|C=\ C\left(x_n\right) \right \}\\
& x_{neg}\in\left \{C| C\neq\  C\left(x_n\right), C\notin C_{imbalance} \right \}\\
& x_{minor}\in \left\{\begin{matrix}
\left\{C|C\neq C\left(x_n\right)\right\} \ if\  |C_{imbalance}|=1 \\
\left\{C|C\neq C\left(x_n\right), C\in C_{imbalance} \right\} else
\end{matrix}\right.
\end{aligned}
\end{equation}

Where $ C(x_n) $ is the class of $ x_n $, $ C_{imbalance} $ is the imbalanced classes subset. The time series data are obtained by applying sliding window method. Set original data matrix  $ \hat{X}=[{\hat{x}}_1;{\hat{x}}_2;{\hat{x}}_3;\ldots;{\hat{x}}_n] \in R^{n\times m} $ where $ n $ is the sample number and $ m $ is the feature number. Suppose the start sampling time is $ t $ and sampling duration is $ L $, then time series data input data can be written as:

\begin{equation}
    x_n=\left[{\hat{x}}_t;{\hat{x}}_{t+1};\ldots;{\hat{x}}_{t+L}\right]
\end{equation}

Sliding window is used between time series samples, and suppose the step length of the sliding window is $ s $:
\begin{equation}
x_{n+1}=\left[{\hat{x}}_{t+s};{\hat{x}}_{t+s+1};\ldots;{\hat{x}}_{t+s+L}\right]
\end{equation}

The label of time series sample is set corresponding to the last sample point in the time series:
\begin{equation}
    y_n=\ {\hat{y}}_{t+L}\ ;\ y_{n+1}={\hat{y}}_{t+s+L}
\end{equation}

\subsection{Feature extract and embedding}
LSTM-QDM utilizes LSTM to make time series feature extraction and a one-layer fully-connected layer to embed the feature and obtain a low-dimensional embedding space. According to the description of LSTM in Section \ref{lstm}, suppose LSTM is $ N_L $ and the extracted feature pair is as follow:
\begin{equation}
N_L\left(Pair_n\right) = 
(N_L\left(x_n\right),N_L\left(x_{pos}\right),N_L\left(x_{neg}\right),N_L(x_{minor}))
\end{equation}

Where $ N_L\left(x\right)\in R^{N_h}$ , means that feature extract layer obtains the last hidden state vector in LSTM. The extracted feature dimension is usually high where Euclidean distance may degenerate and makes it difficult for model to converge at the optimal. So QDM applies a one-layer fully-connected layer to map a high-dimensional space to a low-dimensional space. Suppose the weight of fully-connected layer is $ W_{fe} $ and the mapping vector is $ p $.

\begin{equation}
\begin{aligned}
& p_n=\sigma\left(W_{fe}N_L\left(x_n\right)\right) \\
& p\left(pairs_n\right)=\left(p_n,\ p_{pos},\ p_{neg},\ p_{minor}\right)
\end{aligned}
\end{equation}

The weights of feature extract part (LSTM) and feature embedding part (fully-connected layer) are weight-sharing.

\subsection{Loss function}
\subsubsection{Quadruplet loss design}
\begin{figure}
	\centering
		\includegraphics[scale=.65]{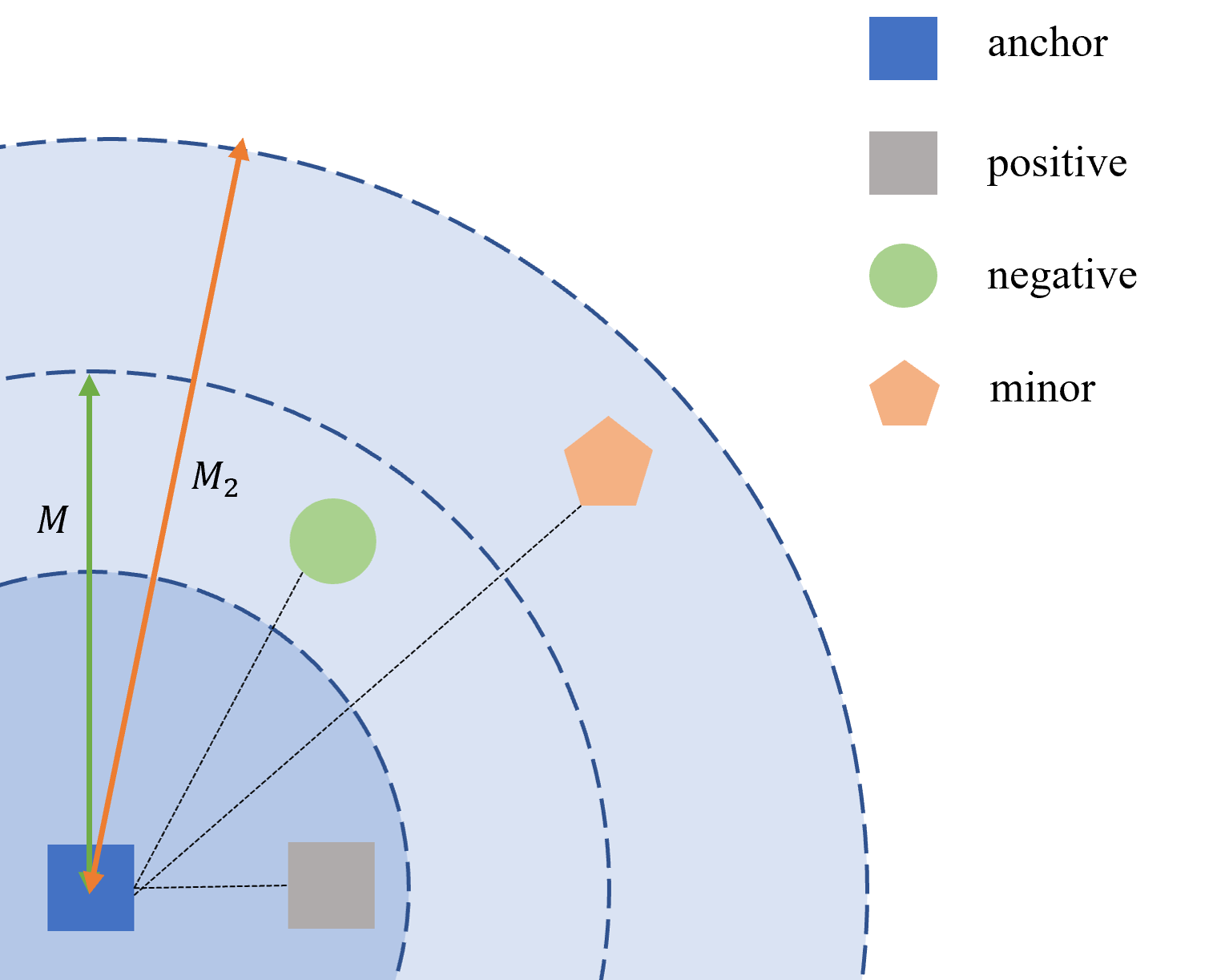}
	\caption{Intuitive interpretation of quadruplet loss}
	\label{img5}
\end{figure}

The design of quadruplet loss refers to contrastive loss. The main intention of quadruplet loss is shown in Figure \ref{img5}. The optimization goal of the proposed loss is to reduce the distance within a class and increase the distance between different classes. Moreover, for imbalanced class, it is ideal that the distance between anchor and imbalanced sample is larger which can improve the performance of imbalanced classification.

Quadruplet loss is composed of three distance from data pair proposed in Section \ref{31}.
\begin{equation}
\begin{aligned}
& D_{n,pos}=\left \| p_n-p_{pos} \right \|_2 \\
& D_{n,neg}=\left \| p_n-p_{neg} \right \|_2 \\
& D_{n,minor}=\left \| p_n-p_{minor} \right \|_2
\end{aligned}
\end{equation}

For sample pair from same class, the distance between samples should be as small as possible which means $ D_{n,pos} $ should be small. Meanwhile, since sample number of balanced class is sufficient, model ought to pay more attention to the data distribution within the balanced class and mine high-quality features from balanced class sample pairs. In such feature space, compact data distribution can improve the negative effect of imbalance. Therefore, a parameter $ \gamma $ is defined which equals 1 if anchor
$ x_n $ belongs to balanced classes and the positive sample part loss is expressed as
\begin{equation}
\begin{aligned}
& \mathcal{L}_{pos}=\left(1-  \gamma \right) D_{n,pos}+ \lambda_{pos} \gamma D_{n,pos}\\
&
\gamma=\left\{\begin{matrix}
0 \ if \ x_n \in C_{imbalance} \\ 
1 \ if \ x_n \notin C_{imbalance}
\end{matrix}\right.
\end{aligned}
\end{equation}

Where $ \lambda_{pos} > 1 $ and is used to emphasize the compactness within balanced class.

For samples from different classes, the distance between samples is hoped to be as large as possible, which means $ D_{n,neg} $ and $ D_{n,minor}$ is large. Moreover, as shown in Figure \ref{img5}, $ D_{n,minor} $ is larger than $ D_{n,neg} $ since farther distance between classes means higher degree of discrimination and reduce the impact of imbalance. Therefore, the loss of negative sample part and minor sample part are as follows:
\begin{equation}
\begin{aligned}
& \mathcal{L}_{neg}=\max\left(0,M-D_{n,neg}\right) \\
& \mathcal{L}_{minor}=\max{\left(0, M_2-D_{n,minor}\right)}\ast \lambda_{minor}\\
\end{aligned}
\end{equation}

Where $ \lambda_{minor}>1 $ and $ M_2 >  M $. $ M $ is the expected margin between anchor and negative sample, and $ M_2 $ is the one between anchor and minor sample. These two formula both pull negative sample and minor sample away from anchor, however, the different margin determines the distance from anchor. For minor sample part loss, the margin is larger so that the distance between anchor and imbalanced class sample is larger. And by adding a hyperparameter $ \lambda_{minor} $, the model pays more attention to the minor data pair to reduce the impact of imbalance.

According to aforementioned three part loss functions and considering a batch input data pairs, the overall  quadruplet loss function is:
\begin{equation}
   \mathcal{L}_{quadruplet}=\frac{1}{N}\sum_{i=1}^N\frac{1}{3}(\ \mathcal{L}_{pos}+\ \mathcal{L}_{neg}+\mathcal{L}_{minor})
\end{equation}

\subsubsection{Softmax loss}
Softmax loss is the most widely used classification loss function in deep learning
\cite{SOFT1, SOFT2} because of its simplicity and excellent performance in many classification tasks and the formula is:

\begin{equation}
    \mathcal{L}_{softmax}=-\frac{1}{N}\sum_{i=1}^{N}y_i\log\frac{e^{a_i}}{\sum_{k=1}^{C}e^{a_k}}
\end{equation}

Where $ y_i $ is the true label of $ i_{th} $ sample, $ a_k $ is the $ k_{th} $ logits corresponding to the $ k_{th} $ class, $ C $ is the class number and $ \frac{e^{a_i}}{\sum_{k=1}^{C}e^{a_k}} $ is the normalized predict probability. In QDM model, the embedding vector of anchor will be mapped into a C-dimension vector by a fully-connected layer. Suppose the weight of layer is $ W_{fc} $, and the mapped vector is
\begin{equation}
    a= \sigma(W_{fc} p_n)
\end{equation}

Because of the existence of $ y_i $ in the formula, the sample number of each class will affect the loss and the model tends to pay attention to the balanced classes and despise the imbalanced classes. Therefore, softmax loss is not suitable for dealing with imbalanced classification problem.

\subsubsection{Combination of quadruplet loss and softmax loss}
From the above introduction of the two losses, softmax loss optimizes the model classification performance, while quadruplet loss optimizes the data distribution in embedding space. Even though the goals of the two losses are different, the optimization of quadruplet loss is positive to softmax loss.
\begin{figure}
	\centering
		\includegraphics[scale=.55]{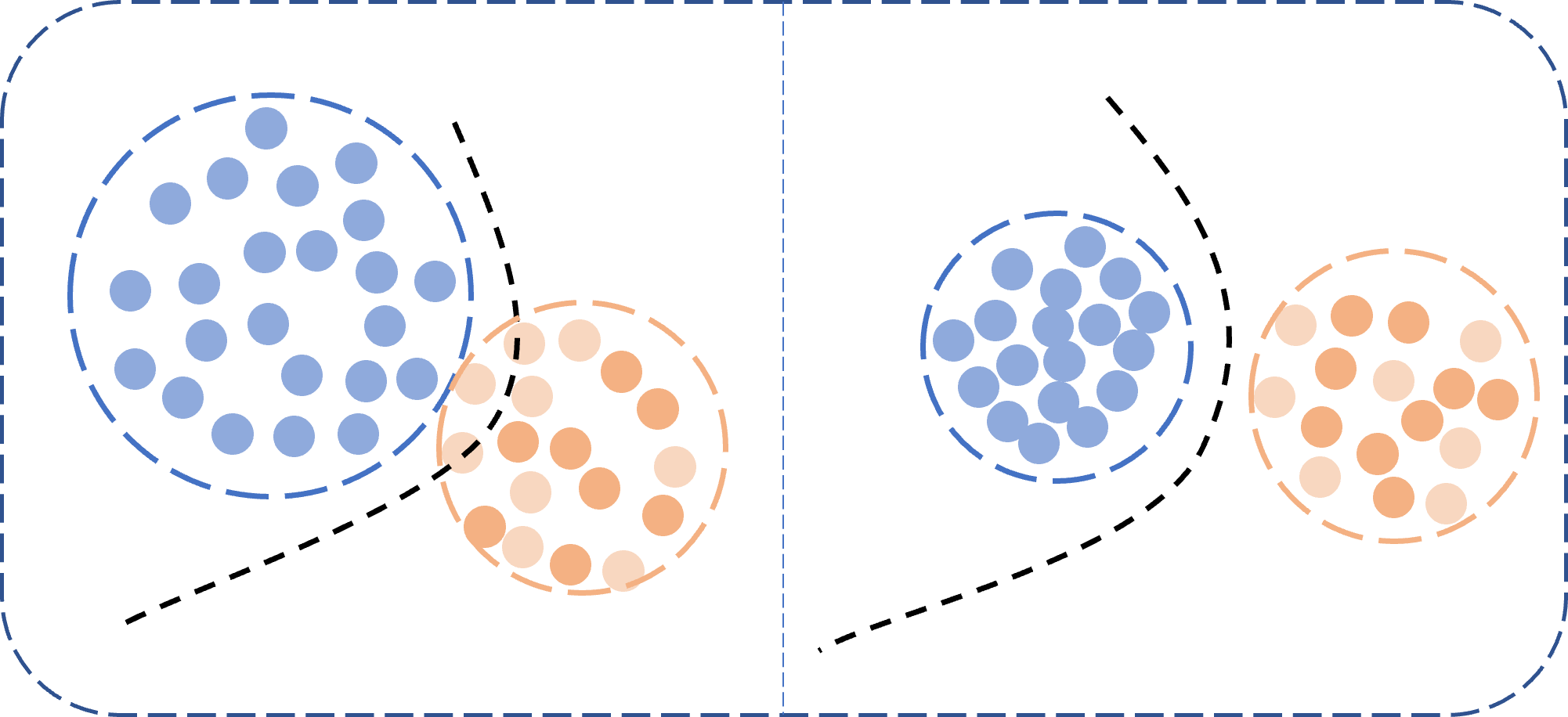}
	\caption{The original imbalanced classification decision boundary(left) and the classification decision boundary after adjusting the data distribution(right)}
	\label{img6}
\end{figure}

As shown in Figure \ref{img6}, the left shows the classification result using softmax loss in the imbalanced classification task and the imbalanced class samples are represented by orange dots. The darker ones are training set and the lighter ones are test set. Because the training set is imbalanced, the model does not fit the decision boundary well and performs unsatisfactorily on the test set. But for the right figure, better performance on both the training set and the test set shows up. Compared with the left, distance between classes in the right figure is larger, and the data distribution of balanced class is more compact. Look back to those two loss functions, with label information $ y_i $ in softmax, the model explicitly considers the number of different class. However, $ y_i $ does not directly appear in quadruplet loss. Moreover, samples belonging to an imbalanced class will structure into data pairs with multiple other classes data which reduces the imbalance influence.

According to above analysis, the loss function of LSTM-QDM combines quadruplet loss and softmax loss with a hyperparameter. The parameters of feature extract and feature embedding part are jointly optimized by these two losses, while the parameters of classifier layer is optimized by softmax loss. The loss function of QDM is as follows.

\begin{equation}
    \mathcal{L}= \mathcal{L}_{softmax}+ \beta\ast \mathcal{L}_{quadruplet}
\end{equation}

\begin{figure*}
	\centering
		\includegraphics[scale=.75]{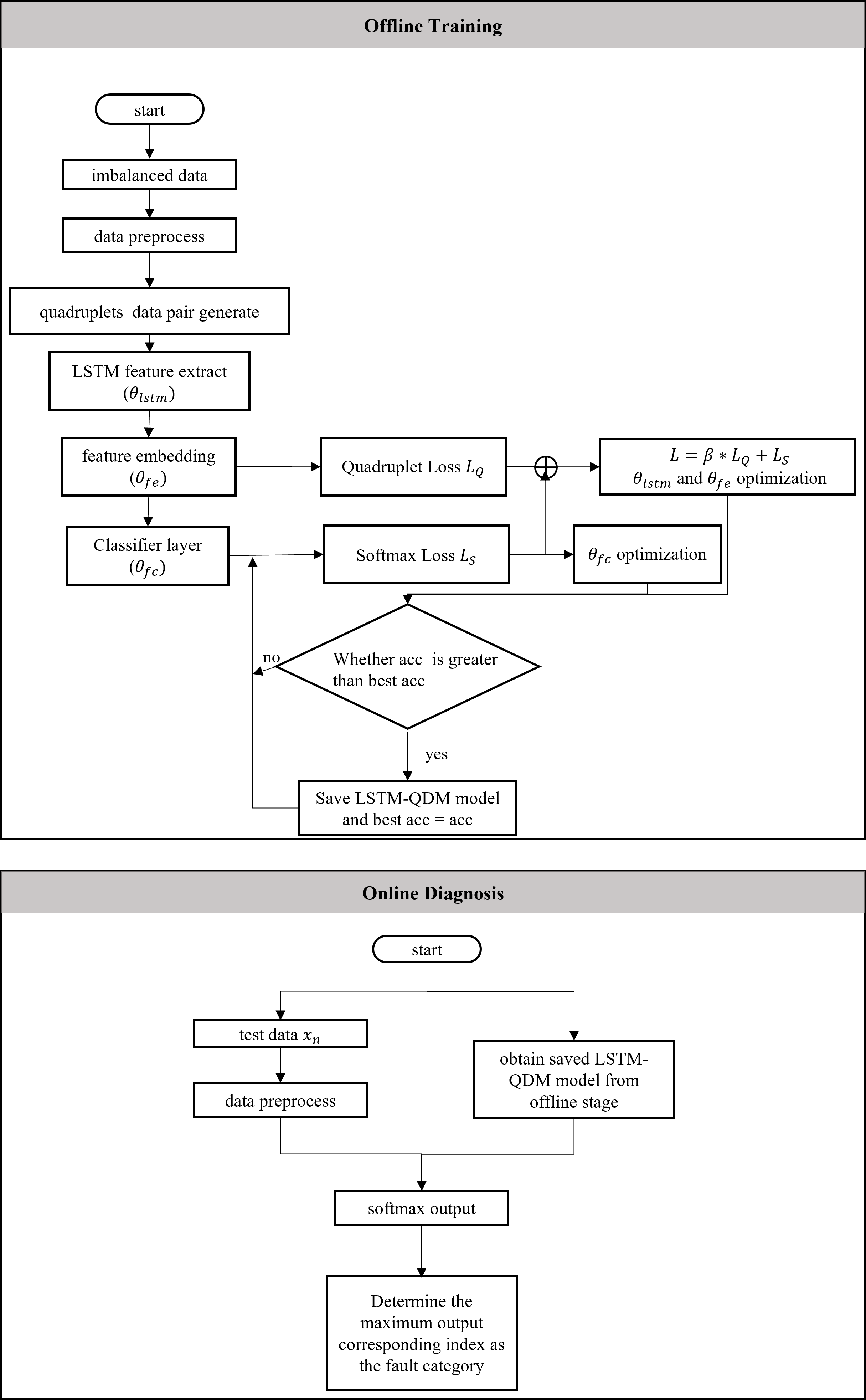}
	\caption{Offline training and online diagnosis of the algorithm.}
	\label{img9}
\end{figure*}
\subsection{Offline Training $ \& $ Online Diagnosis}

This paper proposes an intelligent diagnosis method based on LSTM-QDM to solve the problem of imbalance fault diagnosis. The method is divided into two stages: offline training and online diagnosis. The specific flow chart is shown in Figure \ref{img9}.


\section{Experiment}
To verify the effectiveness of the proposed model, an industrial process fault diagnosis dataset and a bearing fault diagnosis dataset are used to conduct diverse experiments. Detailed experimental analysis will be described in this section.

Two evaluation index, recall rate and F1, are used to evaluate the proposed fault diagnosis model. The fault diagnosis task belongs to multi classification problem and when calculating the evaluation index of class $ i $, class $ i $ is regarded as positive class and other classes as negative class. The specific calculation of the two evaluation indexes are Eq.\ref{Recall} and Eq.\ref{F1} respectively. Recall rate calculates the proportion of the number of correctly predicted positive samples in the total positive sample and $ F1 $ is the comprehensive index of accuracy rate and recall rate which reflecting comprehensive recognition ability and stability of the response model. The higher the two indexes, the better the diagnostic ability of the model. 

\begin{equation}
    Recall\ rate = \frac{TP}{TP + FN}
    \label{Recall}
\end{equation}
\begin{equation}
    F1 = \frac{2TP}{2TP + FN + FP}
    \label{F1}
\end{equation}

\subsection{Case 1: TE process fault dataset}
The first case study is based on Tennessee- Eastman (TE) process data. TE process is a process control case based on actual industrial processes. It first appeared at the American Chemical Society Annual Conference in Chicago in 1990. The TE process was created by Eastman Chemical Company
\cite{TE1} and it has been widely used to research fault diagnosis methods
\cite{TE2,TE3,TE4}.
\subsubsection{Data description}
\begin{figure}
	\centering
		\includegraphics[scale=.45]{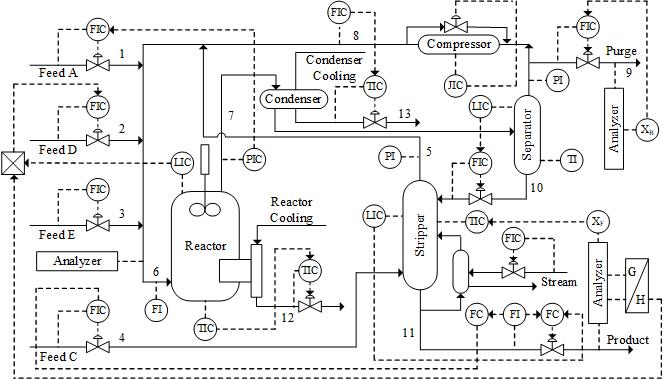}
	\caption{The flowchart of the Tennessee Eastman process}
	\label{tep}
\end{figure}

The TE process consists of five primary units: a reactor, a condenser, a compressor, a separator and a stripper. The whole TE process consists of 22 continuous process measurements, 19 composition measurements and 12 operation variables, and it can generate normal condition and 20 kinds of faults condition. The description of faults is shown in Table \ref{T1} and the flow chart of TE process is shown in Figure \ref{tep}. The data used in this experiment can be downloaded from \url{https://dataverse.harvard.edu/dataset.xhtml?persistentId=doi:10.7910/DVN/6C3JR1}

\begin{table*}[h]
\centering
\caption{Description of 20 faults in TE process}
\label{T1}
\begin{tabular}{cll}
\hline
\textbf{Fault} & \multicolumn{1}{c}{\textbf{Description}} & \multicolumn{1}{c}{\textbf{Type}} \\ \hline
1 & A/C feed ratio, B composition constant   (stream 4) & Step change \\ \cline{1-2}
2 & B composition, A/C ratio constant (stream 4) &  \\ \cline{1-2}
3 & D feed temperature (stream 2) &  \\ \cline{1-2}
4 & Reactor cooling water inlet temperature &  \\ \cline{1-2}
5 & Condenser cooling water inlet temperature &  \\ \cline{1-2}
6 & A feed loss (stream 1) &  \\ \cline{1-2}
7 & C header pressure loss-reduced availability   (stream 4) &  \\ \hline
8 & A, B, C feed composition (stream 4) & Random variation \\ \cline{1-2}
9 & D feed temperature (stream 2) &  \\ \cline{1-2}
10 & C feed temperature (stream 4) &  \\ \cline{1-2}
11 & Reactor cooling water inlet temperature &  \\ \cline{1-2}
12 & Condenser cooling water inlet temperature &  \\ \hline
13 & Reaction kinetics & Slow drift \\ \hline
14 & Reactor cooling water valve & \ {Sticking} \\ \cline{1-2}
15 & Condenser cooling water valve &  \\ \hline
16-20 & Unknown & Unknown \\ \hline
\end{tabular}
\end{table*}

The entire data set includes a normal training set, a normal test set, a fault training set and a fault test set. Each data set contains 500 subdatasets generated in different random states. Each subdataset contains 500 sampling points (training set) or 960 sampling points (test set). For the fault training set, the first 20 sampling points are normal samples and the last 480 sampling points are fault samples, For the fault test set, the first 160 sampling points are normal samples and the last 800 sampling points are fault samples. In order to facilitate the experiment, Fault 1, 5, 6, 8, 12, 16, 20 are selected from the 20 faults for experimental verification. The relevant parameters of the time series data set construction are that the window size is 100, and sliding step size is 1.

\subsubsection{Imbalanced experiments analysis}
To demonstrate the performance of the proposed model for imbalanced classification, several methods are used to compare, including baseline method(LSTM), Siamese network, GAN and Oversample. Note that the neural network architecture and training hyperparameters of those methods are the same as the proposed model which are shown in Table \ref{table2}. Each method is repeat 10 times to measure average evaluation index. Moveover, experiments are conducted on the TE data set under the conditions of single-fault imbalance, multiple-fault imbalance, and different imbalance ratios to prove the robustness of the model.



\begin{table*}[]
\centering
\caption{Parameters of networks in Case 1}
\label{table2}
\begin{tabular}{cll}
\hline
\multicolumn{1}{l}{} & \textbf{Training parameter} & \textbf{Value} \\ \hline
{General hyperparameters} & LSTM hidden size/ layer number /dropout & 100/3/0.5 \\
 & fully-connect   layer & 100*64, 64*7 \\
 & learning rate/   batch size & 1e-3/ 256 \\ \hline
LSTM-QDM & $ M $/ $ M_2 $/ $ \lambda_{pos} $/ $ \lambda_{minor} $/ $ \beta $ & 20/ 50/ 50/ 20/ 5e-4 \\ \hline
LSTM-Siamese & $margin$ / $\beta$ & 20 / 5e-4 \\ \hline
{GAN} & Generator latent dim/ learning rate & 16/ 0.001 \\
 & Discriminator learning rate & 0.01 \\ \hline
\end{tabular}
\end{table*}
Considering single-fault imbalance scenario, fault 8 is selected as imbalanced fault. The number of balanced fault is 4800 and the number of fault 8 is 480 in the training set, and the imbalance ratio is 10:1 (scenario 1). The experiment result is shown in Table \ref{sc1}.

It can be seen from Table1 that the recall rate of fault 8 of the baseline method is 56.07\%, and the F1 score is 71.77\%. The recall rates of methods based on GAN and Oversample are 65.01\%, 64.81\%, with about 9\% increasing, and F1 scores are 78.66\%, 78.38\% with about 7\% increasing, respectively. The recall rate and F1 score of LSTM-Siamese are 61.50\% and 75.95\% respectively, which are only slightly improved compared to the baseline method. The recall rate and F1 score of LSTM-QDM are 66.74\% and 79.85\%, respectively, which is the best performance among all methods. Similarly, the average indicators of proposed method also has best performance and an improvement of about 2\% compared with the baseline method.

To prove the effect of the model on other fault, fault 6 is selected as the imbalanced fault. Similarly, the number of fault 6 is 480 (scenario 2). The result is shown in Table \ref{sc2}. It can be seen that the the recall rate of fault 6 is 80.26\%, and F1 is 87.86\%. Compared with scenario 1, the two evaluation indexes are both higher, indicating that fault 6 is less affected by imbalance. The recall rate of GAN is as high as 93.17\%, with about nearly 13\% increasing compared to the baseline. And the recall rate of LSTM-QDM is 83.60\%, with about nearly 3\% increasing. Even though the performanc of proposed method is not as good as GAN, it is also effective. Similarly, in terms of the average evaluations, the proposed method is lower than GAN, but has a improvement relative to the baseline method.

\begin{table*}[p]
\centering
\caption{Diagnosis performance in scenario 1(fault 8 / 4800:480)}
\label{sc1}
\begin{tabular}{llllll}
\hline
               & LSTM-QDM & LSTM    & LSTM-SIAM & GAN-LSTM & Oversample-LSTM \\ \hline
Fault 8 recall & 66.74\%  & 56.07\% & 61.50\%   & 65.01\%  & 64.81\%         \\
Average recall & 95.00\%  & 93.48\% & 93.95\%   & 94.29\%  & 94.01\%         \\
Fault 8 F1     & 79.85\%  & 71.77\% & 75.95\%   & 78.66\%  & 78.38\%         \\ 
Average F1     & 94.70\%  & 92.83\% & 93.53\%   & 93.91\%  & 93.60\%         \\ \hline
\end{tabular}
\end{table*}

\begin{table*}[p]
\centering
\caption{Diagnosis performance in scenario 2(fault 6 / 4800:480)}
\label{sc2}
\begin{tabular}{llllll}
\hline
               & LSTM-QDM & LSTM    & LSTM-SIAM & GAN-LSTM & Oversample-LSTM \\ \hline
Fault 8 recall & 83.60\%  & 80.26\% & 74.77\%   & 93.17\%  & 83.08\%         \\
Average recall & 97.06\%  & 96.77\% & 95.87\%   & 98.23\%  & 97.17\%         \\
Fault 8 F1     & 90.81\%  & 87.86\% & 85.15\%   & 96.25\%  & 90.46\%         \\ 
Average F1     & 97.00\%  & 96.59\% & 95.70\%   & 97.93\%  & 97.12\%         \\ \hline
\end{tabular}
\end{table*}

Aforementioned two scenarios show that the selected generator and discriminator parameters of GAN are better for fault 6 which means it is greatly affected by different fault scenarios. Although GAN method has a highlight, the shortcomings of GAN is the need of optimizing parameters for different faults. However, The proposed LSTM-QDM has stable improvement compared with baseline method in both scenarios.

To verify the performance of the proposed model under different imbalance ratios, two experiments on fault 8 with imbalance ratio of 5:1 and 20:1(scenario 3 and 4) are conducted where the number of fault 8 is 960 and 240 respectively. The experimental results are shown in Table \ref{sc3} and Table \ref{sc4}.

Tabel \ref{sc3} shows that when the imbalance ratio is 5:1, the recall rate and F1 of baseline method are 63.63\% and 77.66\%, with a increases about 6\% compared with scenario 1 respectively which means that reducing the imbalance ratio can effectively improve the classification performance. Similar to scenario 1, compared with the baseline method, LSTM-QDM improves recall and F1 by 10\% and 7\% respectively, and achieves the best results in all indicators among all methods. 

In scenario 4, the 20:1 imbalance ratio can be regarded as severe imbalance. The recall rate and F1 of imbalance fault 8 of baseline method are only 28.54\% and 44.15\% , which are far lower than scenario 1, indicating such severe imbalance will greatly damage the classification ability of the model. However, although the recall rate and F1 of the proposed LSTM-QDM model are only 52.24\% and 68.46\%, there is still a big improvement compared with the baseline method and similarly, the proposed method has the best performance.

The above four scenarios show that the recall rate and F1 of different imbalanced faults under different imbalance ratios will be greatly affected, as well as the average index of all faults. The method based on GAN and Oversample effectively improves the performance by re-balancing the number of different classes. The method based on LSTM-Siamese has a certain improvement compared to the baseline, because it can improve the sample distribution and class distance in the mapping space to a certain extent. However, the model does not pay special attention to the imbalance class, so the overall effect is worse than the re-balancing method. The proposed LSTM-QDM model deliberately adds a data pair of anchor and minor sample. For all input quadruplet data pairs, the distance between imbalanced sample and other samples is modeled, and utilizing hyperparameters, $ M_1 $ ,$ M_2 $ , $ \lambda_{pos} $, $ \lambda_{minor} $  to better fit the intra-class distribution and inter-class distance so the performance is greatly improved.
\begin{table*}[p]
\centering
\caption{Diagnosis performance in scenario 3(fault 8 / 4800:960)}
\label{sc3}
\begin{tabular}{llllll}
\hline
               & LSTM-QDM & LSTM    & LSTM-SIAM & GAN-LSTM & Oversample-LSTM \\ \hline
Fault 8 recall & 73.64\%  & 63.63\% & 67.48\%   & 68.43\%  & 66.69\%         \\
Average recall & 96.11\%  & 94.73\% & 95.35\%   & 95.43\%  & 95.17\%         \\
Fault 8 F1     & 84.68\%  & 77.66\% & 80.44\%   & 81.11\%  & 79.93\%         \\ 
Average F1     & 95.93\%  & 94.43\% & 95.03\%   & 95.13\%  & 94.89\%         \\ \hline
\end{tabular}
\end{table*}

\begin{table*}[p]
\centering
\caption{Diagnosis performance in scenario 4(fault 8 / 4800:240)}
\label{sc4}
\begin{tabular}{llllll}
\hline
               & LSTM-QDM & LSTM    & LSTM-SIAM & GAN-LSTM & Oversample-LSTM \\ \hline
Fault 8 recall & 52.24\%  & 28.54\% & 45.39\%   & 48.11\%  & 47.91\%         \\
Average recall & 92.78\%  & 89.40\% & 91.86\%   & 92.35\%  & 92.45\%         \\
Fault 8 F1     & 68.46\%  & 44.15\% & 62.09\%   & 64.65\%  & 64.41\%         \\ 
Average F1     & 92.05\%  & 87.19\% & 90.77\%   & 92.06\%  & 91.84\%         \\ \hline
\end{tabular}
\end{table*}

When there are multiple faults in imbalanced scenario, faults 6, 8, and 16 are considered as imbalanced classes. Similarly, considering two imbalance ratios, 10:1 and 5:1. The number of balanced faults is 4800 and the number of imbalanced faults are 480, 960 respectively. In addition, in these two experiments, the baseline method performance under balanced conditions are also shown. The experiment results are shown in Table \ref{multi1} and Table \ref{multi2}.

The rightmost column of the two tables shows the result of LSTM method under balanced conditions. It can be seen that all index are close to 100\%, indicating that the settings of the general hyperparameters in Table \ref{table2} are reasonable. From Table \ref{multi1}, it can be seen that the recall rates of fault 6, fault 8 and fault 16 of baseline method are 73.04\%, 46.59\%, and 87.12\% respectively, and the F1 scores are 83.01\%, 54.91\%, and 88.67\% respectively. This shows that fault 8 is the most severely affected class by imbalance, fault 6 is the second, and fault 16 is the least. Compared with the baseline method, the recall rate of these three imbalanced fault of LSTM-QDM are with about nearly 6\%, 16\%, 10\% increase and the F1 are with about 5\%, 19\% and 4\% increase respectively and have best performance in all evaluation indicators, including fault average index, except for fault 6 where the result of GAN far exceeds other methods.The performance of GAN method in this scenario and scenario 2 is consistent, which shows that the current hyperparameters of GAN are suitable for fault 6.However, in terms of overall performance, LSTM-QDM is stable.

Table 8 shows the results of scenario 6 where the imbalance ratio reaches 5:1. It can be seen that almost all indexes are improved compared to Table \ref{multi1}, except for the GAN method on fault 6, which indicating that GAN is unstable and more complex data may make optimization performance of GAN not ideal under current hyperparameters. The recall rate of LSTM-QDM on fault 6 and fault 16 reach 97.36\% and 99.38\%, and the F1 score reach 98.14\% and 93.63\%, which are close to the performance under the balanced scenario. Compared with the baseline method, the performance of fault 8 is also improved by nearly 10\%. In terms of overall average performance, LSTM-QDM also has an excellent performance.

In the above experiments of TE dataset, the diverse scenarios include different fault classes, imbalanced ratios, and single-fault or multi-fault imbalance are conducted for comparison. The experimental results show that the proposed LSTM-QDM method has good performance under different experimental conditions, which proves the effectiveness and robustness of the model on this dataset.

\begin{table*}[p]
\centering
\caption{Diagnosis performance in scenario 5(fault 6,8,16 / 4800:480)}
\label{multi1}
\begin{tabular}{lllllll}
\hline
                & LSTM-QDM & LSTM    & LSTM-SIAM & GAN-LSTM & Oversample-LSTM & Balanced-LSTM \\ \hline
Fault 6 recall  & 79.68\%  & 73.04\% & 76.24\%   & 95.36\%  & 71.88\%         & 99.97\%       \\
Fault 6 F1      & 88.14\%  & 83.01\% & 83.36\%   & 96.42\%  & 83.30\%         & 99.98\%       \\
Fault 8 recall  & 62.28\%  & 46.59\% & 49.27\%   & 59.37\%  & 51.41\%         & 98.59\%       \\
Fault 8 F1      & 75.04\%  & 54.91\% & 64.30\%   & 71.69\%  & 64.88\%         & 99.28\%       \\
Fault 16 recall & 97.91\%  & 87.12\% & 95.01\%   & 81.21\%  & 85.35\%         & 100\%         \\
Fault 16 F1     & 92.53\%  & 88.67\% & 92.31\%   & 87.58\%  & 85.27\%         & 99.46\%       \\
Average recall  & 91.37\%  & 86.62\% & 88.63\%   & 90.65\%  & 86.67\%         & 99.79\%       \\ 
Average F1      & 90.69\%  & 85.72\% & 87.75\%   & 90.08\%  & 85.72\%         & 99.76\%       \\ \hline
\end{tabular}
\end{table*}

\begin{table*}[p]
\centering
\caption{Diagnosis performance in scenario 6(fault 6,8,16 / 4800:960)}
\label{multi2}
\begin{tabular}{lllllll}
\hline
                & LSTM-QDM & LSTM    & LSTM-SIAM & GAN-LSTM & Oversample-LSTM & Balanced-LSTM \\ \hline
Fault 6 recall  & 97.36\%  & 88.76\% & 94.78\%   & 90.82\%  & 88.48\%         & 99.97\%       \\
Fault 6 F1      & 98.14\%  & 93.28\% & 96.81\%   & 93.67\%  & 93.04\%         & 99.98\%       \\
Fault 8 recall  & 69.21\%  & 59.59\% & 65.28\%   & 66.88\%  & 65.96\%         & 98.59\%       \\
Fault 8 F1      & 81.64\%  & 71.85\% & 78.34\%   & 78.81\%  & 78.48\%         & 99.28\%       \\
Fault 16 recall & 99.38\%  & 94.49\% & 96.86\%   & 96.30\%  & 94.87\%         & 100\%         \\
Fault 16 F1     & 93.63\%  & 92.64\% & 92.22\%   & 92.73\%  & 93.79\%         & 99.46\%       \\
Average recall  & 95.04\%  & 91.78\% & 93.80\%   & 93.33\%  & 92.57\%         & 99.79\%       \\ 
Average F1      & 94.81\%  & 91.37\% & 93.51\%   & 93.02\%  & 92.31\%         & 99.76\%       \\ \hline
\end{tabular}
\end{table*}

\subsection{Case 2: CWRU bearing fault dataset}
The second case study is based on CWRU bearing data. CWRU bearing dataset was collected from an experiment platform provided by Case Western Reserve University and has been widely used in the field of mechanical fault detection and fault diagnosis.\cite{CWRU1,CWRU2,CWRU3}

\subsubsection{Data description}
On the CWRU bearing experiment platform, experiments were conducted using an electric motor, and vibration data were measured from the motor bearings. The main components of this experimental facility consist of a 2 hp motor, a torque transduce and a dynamometer. The test bearing support the motor shaft. The rolling bearing is tested under different loads (0, 1, and 2 HP). The location of the fault mainly includes ball defect (BD), outer race defect (OR) and inner race defect (IR) and the defect diameter includes 0.007, 0.014, 0.021 inches. The vibration signals are collected by the accelerometer sensor with 12 kHz.

The collected accelerometer signals of the drive end under 0 HP loads are used for method validation in this case which is a one-dimensional time-series data. There are one class of normal data and nine classes of fault data. Each type of fault uses a sliding window method to construct time series data. The window size used here is 400, and the sliding step size is 32. The rolling bearings fault information are shown in Table \ref{CWRU}.

\begin{table}[h]
\centering
\caption{Description of faults in CWRU}
\label{CWRU}
\begin{tabular}{lll}
\hline
\multicolumn{1}{c}{\textbf{Fault location}} & \multicolumn{1}{c}{\textbf{Fault diameter}} & \multicolumn{1}{c}{\textbf{Label}} \\ \hline
Normal     & 0     & 0 \\
Ball       & 0.007 & 1 \\
Ball       & 0.014 & 2 \\
Ball       & 0.022 & 3 \\
Inner race & 0.007 & 4 \\
Inner race & 0.014 & 5 \\
Inner race & 0.022 & 6 \\
Outer race & 0.007 & 7 \\
Outer race & 0.014 & 8 \\
Outer race & 0.022 & 9 \\ \hline
\end{tabular}
\end{table}

\subsubsection{Imbalanced experiments analysis}
Similar to case 1, the performance of aforementioned five methods are compared under different situations in this case. The hyperparameters of the model have been adjusted and are shown in the Table \ref{paramcase2}. Each method is also repeated ten times and the average performance is taken as the result. The number of samples for each class is 1800 in the training set and 200 in the test set and the number of imbalance classes in the training set is adjusted according to different imbalance ratios. Fault 3, 5, and 9 are used as imbalanced classes for experiments which come from different fault locations.

\begin{table*}[]
\centering
\caption{Parameters of networks in Case 2}
\label{paramcase2}
\begin{tabular}{cll}
\hline
\multicolumn{1}{l}{} & \textbf{Training parameter} & \textbf{Value} \\ \hline
{General hyperparameters} & LSTM hidden size/ layer number /dropout & 30/3/0.1 \\
 & fully-connect   layer & 30*15, 15*10 \\
 & learning rate/   batch size & 5e-2/ 128 \\ \hline
LSTM-QDM & $ M $/ $ M_2 $/ $ \lambda_{pos} $/ $ \lambda_{minor} $/ $ \beta $ & 5/ 10/ 10/ 10/ 1e-3 \\ \hline
LSTM-Siamese & $margin$ / $\beta$ & 5 / 1e-3 \\ \hline
{GAN} & Generator latent dim/ learning rate & 20/ 0.001 \\
 & Discriminator learning rate & 0.01 \\ \hline
\end{tabular}
\end{table*}

Considering 10:1 imbalance ratio and the imbalance class is fault 3 as scenario 1 which means the number of fault 3 in training set is 180. The result is shown in Table \ref{CWRU1}. It can be seen that the recall rate and F1 of fault 3 of the baseline method are only 11.23\% and 18.22\% respectively, which are greatly affected by the imbalance. The two indexs of LSTM-QDM are 48.05\% and 63.70\% with about 37\% and 45\% increase respectively. Among all the methods, GAN has the best effect, and the two indexes of which are about 4\% larger than LSTM-QDM. Under the current parameter settings, the re-balancing effect of GAN on imbalance is better than the proposed method. However, LSTM-QDM still improves the impact of imbalance to a certain extent.

\begin{table*}[p]
\centering
\caption{Diagnosis performance in scenario 1(fault 3 / 1800:180)}
\label{CWRU1}
\begin{tabular}{llllll}
\hline
               & LSTM-QDM & LSTM    & LSTM-SIAM & GAN-LSTM & Oversample-LSTM \\ \hline
Fault 3 recall & 48.05\%  & 11.23\% & 43.60\%   & 52.72\%  & 45.05\%         \\
Average recall & 94.18\%  & 90.03\% & 93.97\%   & 94.95\%  & 93.97\%         \\
Fault 3 F1     & 63.70\%  & 18.22\% & 59.17\%   & 67.02\%  & 59.17\%         \\ 
Average F1     & 93.66\%  & 87.13\% & 93.24\%   & 94.40\%  & 92.748\%         \\ \hline
\end{tabular}
\end{table*}

Considering the performance of proposed model on other faults, the imbalanced class is adjusted to fault 5 and fault 9 for experiments in scenario 2 and scenario 3. The experimental results are shown in Table \ref{CWRU2} and \ref{CWRU9}. The recall rates of baseline on imbalanced faults are 90.68\%, 90.22\%, and F1 are 93.76\%, 93.57\% respectively, indicating that the impact of imbalance is relatively slight. And the recall rates of LSTM-QDM on unbalanced faults are 95.64\%, 96.00\%, and F1 are 97.32\% , 98.91\% with about 5\% increasing compared with baseline method. But in the two scenarios, re-balancing method still has best performance.

\begin{table*}[p]
\centering
\caption{Diagnosis performance in scenario 2(fault 5 / 1800:180)}
\label{CWRU2}
\begin{tabular}{llllll}
\hline
               & LSTM-QDM & LSTM    & LSTM-SIAM & GAN-LSTM & Oversample-LSTM \\ \hline
Fault 3 recall & 95.64\%  & 90.68\% & 90.80\%   & 98.55\%  & 99.70\%         \\
Average recall & 98.75\%  & 96.46\% & 96.74\%   & 99.47\%  & 99.02\%         \\
Fault 3 F1     & 97.32\%  & 93.76\% & 93.62\%   & 99.04\%  & 99.55\%         \\ 
Average F1     & 98.75\%  & 96.07\% & 96.71\%   & 99.47\%  & 99.00\%         \\ \hline
\end{tabular}
\end{table*}

\begin{table*}[p]
\centering
\caption{Diagnosis performance in scenario 3(fault 9 / 1800:180)}
\label{CWRU3}
\begin{tabular}{llllll}
\hline
               & LSTM-QDM & LSTM    & LSTM-SIAM & GAN-LSTM & Oversample-LSTM \\ \hline
Fault 3 recall & 96.00\%  & 90.22\% & 94.31\%   & 95.94\%  & 96.70\%         \\
Average recall & 98.83\%  & 97.57\% & 98.89\%   & 99.20\%  & 98.09\%         \\
Fault 3 F1     & 97.60\%  & 93.57\% & 96.77\%   & 97.64\%  & 98.49\%         \\ 
Average F1     & 98.91\%  & 97.81\% & 98.84\%   & 99.16\%  & 98.02\%         \\ \hline
\end{tabular}
\end{table*}

In order to verify the effectiveness of LSTM-QDM under different imbalance ratios, imbalance ratio of different faults is adjusted. Because fault 3 is seriously affected by imbalance, while fault 5 and fault 9 are slightly affected by imbalance, the imbalance ratio of fault 3 is adjusted to 5:1, and the imbalance ratio of fault 5 and fault 9 is adjusted to 20:1. The experimental results are shown in Table \ref{CWRU3},\ref{CWRU4},\ref{CWRU5}. 

By reducing the imbalance ratio of fault 3, it can be seen that the performance of the baseline method has been significantly improved. In this scenario, the recall rate and F1 of LSTM-QDM have increased by about 14\% and 18\% compared with baseline respectively. It can be seen from Table \ref{CWRU4} and Table \ref{CWRU5} that the increase of imbalance ratio will destroy the classification performance, however, even if the imbalance ratio reaches 20:1, the recall rate of baseline method of fault 5 and fault 9 are about 86\% indicating the great distinction between these two faults and other faults. Similarly, the proposed method has a obvious improvement in all indicators.

\begin{table*}[p]
\centering
\caption{Diagnosis performance in scenario 4(fault 3 / 1800:360)}
\label{CWRU3}
\begin{tabular}{llllll}
\hline
               & LSTM-QDM & LSTM    & LSTM-SIAM & GAN-LSTM & Oversample-LSTM \\ \hline
Fault 3 recall & 79.38\%  & 65.81\% & 70.31\%   & 89.11\%  & 85.80\%         \\
Average recall & 97.06\%  & 90.02\% & 96.36\%   & 98.47\%  & 97.43\%         \\
Fault 3 F1     & 85.70\%  & 67.68\% & 79.10\%   & 81.53\%  & 88.73\%         \\ 
Average F1     & 96.96\%  & 89.40\% & 96.10\%   & 98.41\%  & 97.43\%         \\ \hline
\end{tabular}
\end{table*}

\begin{table*}[p]
\centering
\caption{Diagnosis performance in scenario 5(fault 5 / 1800:90)}
\label{CWRU4}
\begin{tabular}{llllll}
\hline
               & LSTM-QDM & LSTM    & LSTM-SIAM & GAN-LSTM & Oversample-LSTM \\ \hline
Fault 3 recall & 94.83\%  & 86.42\% & 91.50\%   & 97.30\%  & 99.25\%         \\
Average recall & 98.71\%  & 96.89\% & 98.01\%   & 98.46\%  & 99.31\%         \\
Fault 3 F1     & 96.68\%  & 91.96\% & 95.06\%   & 98.36\%  & 99.61\%         \\ 
Average F1     & 98.72\%  & 96.87\% & 98.04\%   & 98.42\%  & 99.31\%         \\ \hline
\end{tabular}
\end{table*}

\begin{table*}[p]
\centering
\caption{Diagnosis performance in scenario 6(fault 9 / 1800:90)}
\label{CWRU5}
\begin{tabular}{llllll}
\hline
               & LSTM-QDM & LSTM    & LSTM-SIAM & GAN-LSTM & Oversample-LSTM \\ \hline
Fault 3 recall & 92.94\%  & 86.45\% & 89.93\%   & 90.93\%  & 93.56\%         \\
Average recall & 98.89\%  & 97.61\% & 98.40\%   & 98.43\%  & 98.23\%         \\
Fault 3 F1     & 96.33\%  & 92.62\% & 94.63\%   & 95.07\%  & 96.60\%         \\ 
Average F1     & 98.88\%  & 97.56\% & 98.37\%   & 98.42\%  & 98.18\%         \\ \hline
\end{tabular}
\end{table*}

In order to verify the performance of LSTM-QDM under multiple imbalanced faults situation, faults 3, 6, 9 are selected as imbalanced classes, with 10:1 imbalance ratio. The experimental results are shown in Table \ref{multi3}. Similarly, the rightmost column of the table shows the performance under balanced conditions, and the results show that the general parameter settings are reasonable.

Results of fault 3 and fault 9 in the baseline method are similar to those of the previous single fault experiment. Fault 6 and fault 5 belong to the inner race fault, and the result of fault 6 is similar to that of the fault 5. LSTM-QDM method has a significant improvement in the indicators of the three faults and average indicators. The result shows that the model has stable performance in the multi-fault imbalanced scenario.

From the above seven scenarios, even if the overall performance of the re-balancing method is better than that of LSTM-QDM, LSTM-QDM still shows a relatively good ability to improve classification performance under imbalance. Combined with 6 different experimental settings in Case 1, these experimental results show that LSTM-QDM effectively improves the classification effect of imbalanced classes and overall performance in imbalanced multi-classification tasks.

\begin{table*}[]
\centering
\caption{Diagnosis performance in scenario 7(fault 3,6,9 / 1800:180)}
\label{multi3}
\begin{tabular}{lllllll}
\hline
                & LSTM-QDM & LSTM    & LSTM-SIAM & GAN-LSTM & Oversample-LSTM & Balanced-LSTM \\ \hline
Fault 3 recall  & 42.19\%  & 17.28\% & 27.12\%   & 44.28\%  & 37.45\%         & 99.67\%       \\
Fault 3 F1      & 58.43\%  & 26.77\% & 41.37\%   & 58.70\%  & 51.75\%         & 99.83\%       \\
Fault 6 recall  & 95.31\%  & 91.77\% & 96.06\%   & 97.22\%  & 98.15\%         & 99.33\%       \\
Fault 6 F1      & 96.61\%  & 93.40\% & 96.59\%   & 97.41\%  & 98.84\%         & 99.63\%       \\
Fault 9 recall  & 93.40\%  & 88.11\% & 88.25\%   & 93.72\%  & 96.50\%         & 99.10\%         \\
Fault 9 F1      & 95.29\%  & 91.16\% & 92.50\%   & 94.05\%  & 98.13\%         & 99.66\%       \\
Average recall  & 92.78\%  & 88.64\% & 90.89\%   & 93.25\%  & 91.52\%         & 99.49\%       \\ 
Average F1      & 92.03\%  & 86.35\% & 89.37\%   & 92.39\%  & 90.71\%         & 99.61\%       \\ \hline
\end{tabular}
\end{table*}

\subsection{Analysis of the hyperparameters in LSTM-QDM}
In this section, some analysis and discussion on some hyperparameters in the LSTM-QDM model are conducted including the parameters in quadruplet loss and the weight of the quadruplet loss. 
\subsubsection{Analysis of the hyperparameters in quadruplet loss}
\begin{figure}[h]
	\centering
		\includegraphics[scale=.65]{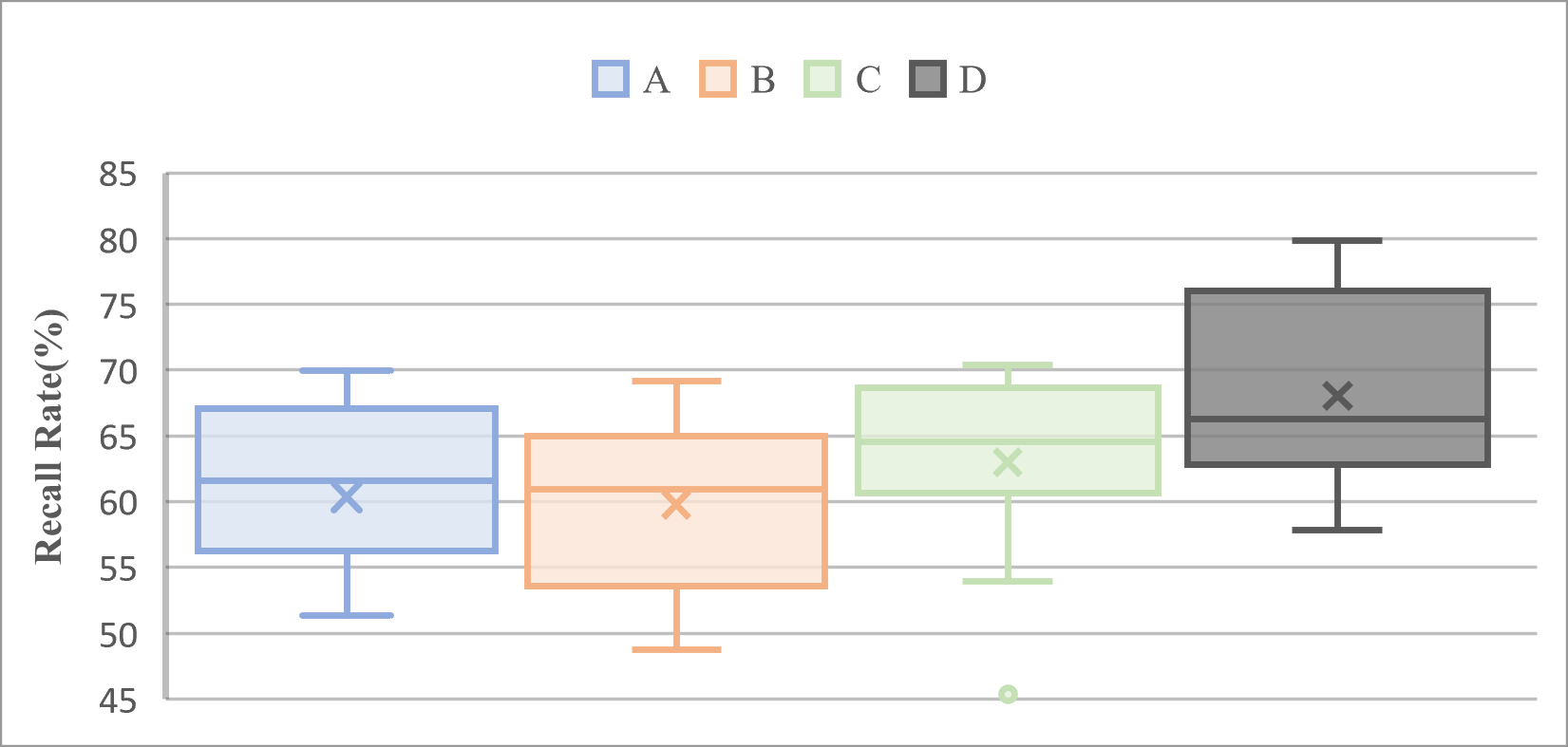}
	\caption{Box plot of recall rates for four parameter settings}
	\label{box}
\end{figure}

Looking back at the quadruplet loss, the hyperparameters include a margin $ M_2 $ larger than $ M $ in traditional siamese network which is used to push imbalance sample further, and the weight parameter $ \lambda_{pos} $ , $ \lambda_{minor} $ that provides different weights for different sample pairs. Different parameters in scenario 1 experiment in the TE data set are set, which are (A) $ M_2 = M = 20, \lambda_{pos} = 1, \lambda_{minor} = 1 $. (B) $ M_2 = M = 20, \lambda_{pos} = 50, \lambda_{minor} = 20 $. (C) $ M_2 = 50, \lambda_{pos} = 1, \lambda_{minor} = 1 $. (D) $ M_2 = 50, \lambda_{pos} = 50, \lambda_{minor} = 20 $. The parameter setting refers to case 1 and each experiment is also conducted ten times. The experimental results are shown in the Figure \ref{box}.

(A) only considers the addition of an imbalanced minor sample pair. (B) does not consider $ M_2 $ and only the weights of different sample pairs are considered. The imbalanced class recall rate decreases slightly, because when all sample margins are the same, it will be more difficult to converge when paying special attention to negative sample pairs containing imbalanced classes. (C) does not consider the weight and only $ M_2 $ is considered, which means the imbalanced class in the embedding space is farther from other classes, and the recall rate of the imbalanced class is improved. When considering $ M_2 $ and weight simultaneously which is (D), the model has the best results, indicating that more attention paid to minor sample pairs and balanced positive samples makes the distribution of balanced data more compact and the distance between classes larger. It should be noted that these parameters need to be adjusted according to different data conditions like the different settings in Case 1 and Case 2.

\subsubsection{Analysis of the hyperparameter $ \beta $}

\begin{figure}[h]
	\centering
		\includegraphics[scale=.58]{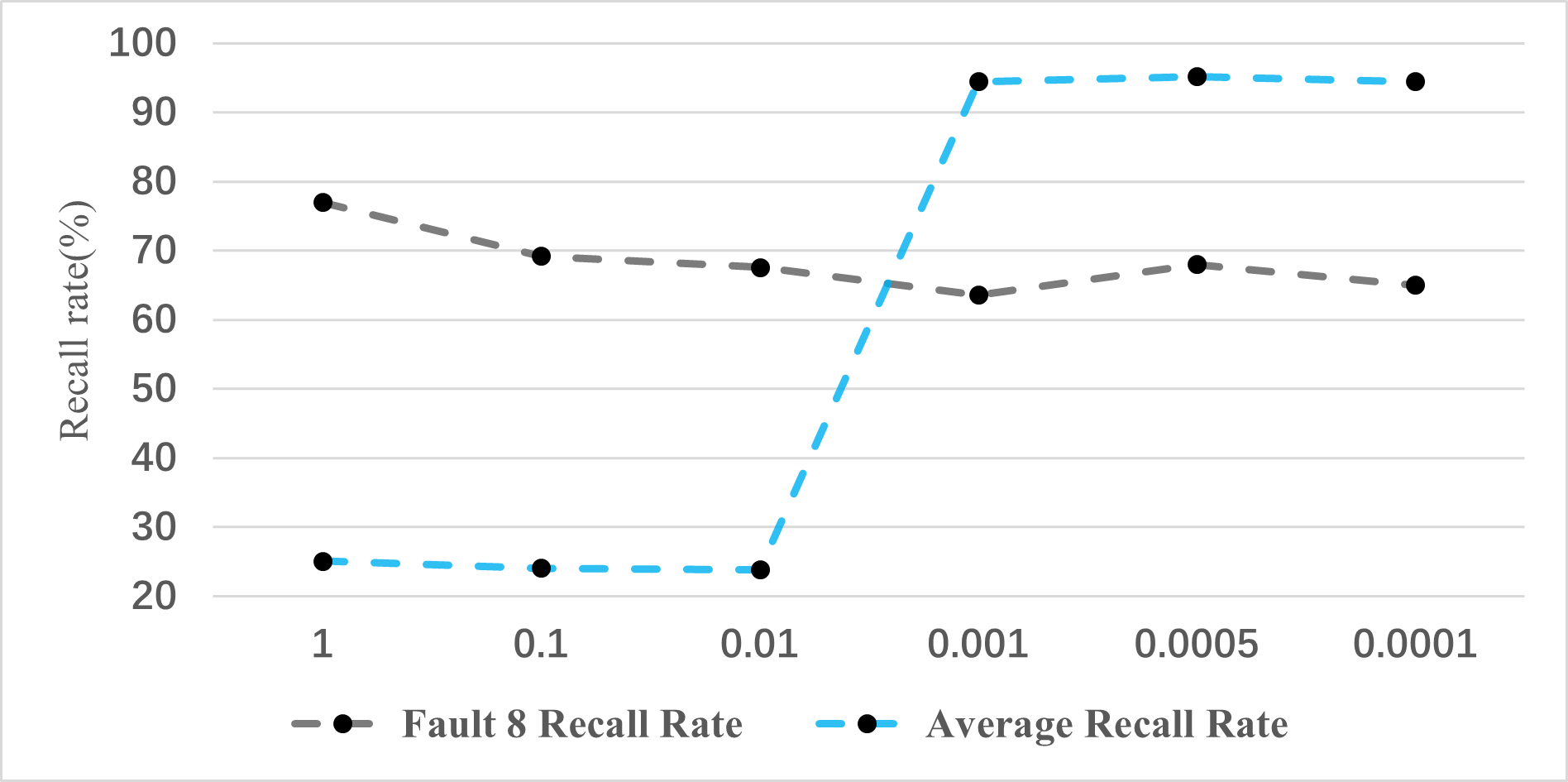}
	\caption{The recall rate of fault 8 and average}
	\label{line}
\end{figure}

The loss function of LSTM-QDM model is the weighted summation of softmax and quadruplet loss, where the parameter $ \beta $ controls the weight of quadruplet loss. Similarly, in this section, through the expansion of scenario 1 experiment in TE data set, parameter $ \beta $ is adjusted and the result is shown in Figure \ref{line}. 

The experimental results show that the $ \beta $ greatly affects the performance of LSTM-QDM. When $ \beta $ is large, although the model has a high recall rate of imbalanced faults, the overall average recall rate is extremely low, indicating that the proportion of quadrupelet loss is too large and the attention on imbalanced class is too high. As a result, the model can not fit other classes well and have unsatisfactory overall performance. However, when $ \beta $ is small, although the recall rate of imbalanced faults is slightly reduced, the average recall rate reaches a higher level which is an ideal phenomenon and indicating that a good balance is achieved between the two losses. Therefore, in the previous experiments, the hyperparameter $ \beta $ was adjusted between 0.001 and 0.0001. 

\section{Conclusion}
In this paper, a time series fault diagnosis model is proposed based on deep metric learning for class imbalance in fault diagnosis task. The proposed model designs a new data pair referring to traditional deep metric method especially considering the imbalanced fault class and proposes a quadruplet loss function. Combined with softmax loss function, the distance between classes and the data distribution within a class are adjusted to improve the imbalance classification performance.

The method proposed in this paper considers multiple-class imbalance and one-class imbalance scenarios simultaneously and designs different data pair construction methods. A large number of experiments on two public datasets (TE, CWRU) are conducted to prove the effectiveness and robustness of the model, and discusses some hyperparameters of the model in depth. This method provides a new perspective to solve the problem of unbalanced classification and further research on deep metric learning in imbalanced classification tasks will be interesting.

\bibliographystyle{unsrt}  
\bibliography{references}  






\end{document}